\documentclass[final]{cvpr}

\usepackage{epsfig}
\usepackage{graphicx}
\usepackage{amsmath}
\usepackage{amssymb}

\usepackage{comment}
\usepackage{subfigure}
\usepackage{color}
\usepackage{url}
\usepackage{algorithm}
\usepackage{url}
\usepackage{algorithmic}
\usepackage{multirow}
\usepackage{smil}
\usepackage{arydshln}
\newcommand*\samethanks[1][\value{footnote}]{\footnotemark[#1]}

\usepackage[margin=4pt,font=small,labelfont=bf,labelsep=endash,tableposition=top]{caption}

\usepackage[pagebackref=true,breaklinks=true,colorlinks,bookmarks=false]{hyperref}

\def\cvprPaperID{2372} %
\def\confYear{CVPR 2021}

\definecolor{mypink}{cmyk}{0, 0.7808, 0.4429, 0.1412}

\begin{document}

\title{AQD: Towards Accurate Fully-Quantized Object Detection}

\author{Peng Chen$^2$\thanks{
First two authors contributed equally.} ~~ ~ Jing Liu$^1$\samethanks ~~ ~ Bohan Zhuang$^1$\thanks{
Corresponding author. Email: $\tt  bohan.zhuang@gmail.com$}  ~~ ~
Mingkui Tan$^{3}$  ~~ ~ Chunhua Shen$^{1,2}$
\\[0.152cm]
	$^1$Monash University ~~ ~
	$^2$The University of Adelaide ~~ ~
	$^3$South China University of Technology
}

\maketitle
\pagestyle{empty}
\thispagestyle{empty}

\begin{abstract}
Network quantization allows inference to be conducted using low-precision arithmetic for improved inference efficiency of deep neural networks on edge devices. However, designing aggressively low-bit (e.g., 2-bit) quantization schemes on complex tasks, such as object detection, still remains challenging in terms of 
severe performance degradation and 
unverifiable efficiency on common hardware. 
In this paper, we propose an Accurate Quantized object Detection solution, termed AQD, to fully get rid of floating-point computation.
To this end, we target using fixed-point 
operations in all kinds of layers, including the convolutional layers, normalization layers, and skip connections, allowing the inference to be executed using integer-only arithmetic. To demonstrate the improved latency-vs-accuracy trade-off, we apply the proposed methods on RetinaNet and FCOS. In particular, experimental results on MS-COCO dataset show that our AQD achieves comparable or even better performance compared with the full-precision counterpart under extremely low-bit schemes, which is of great practical value. 
Source code and %
models are available at:  
 \def\UrlFont{\sf}
    \def\UrlFont{\rm\small\ttfamily}
\url{https://github.com/aim-uofa/model-quantization}
\end{abstract}

\section{Introduction}
\label{sec:introduction}
Deep neural networks (DNNs) have achieved great success in many computer vision tasks, such as image classification~\cite{he2016deep,krizhevsky2012imagenet,he2016identity}, semantic segmentation~\cite{long2015fully,he2017mask,chen2017deeplab}, object detection~\cite{ren2015faster, lin2017focal, lin2017feature}, etc. However, DNNs are always equipped with a large number of parameters and consume heavy computational resources, which hinders their applications, especially on resource-constrained devices such as smartphones and drones. To reduce the memory footprint and computational burden, several network compression methods have been proposed, such as channel pruning~\cite{he2017channel, luo2017thinet, zhuang2018discrimination}, efficient architecture design~\cite{howard2017mobilenets, sandler2018inverted, pham2018efficient} and network quantization~\cite{zhou2016dorefa, zhuang2018towards, zhuang2018towards}. 

\begin{table}[t]
\begin{center}
\caption{Energy and Area cost for different precision operations (on 45nm CMOS technology) \cite{EIE2016Han, isscc2014hard, Lian2019High}. }
\label{table:hardare_cost}
\scalebox{0.9}{
\begin{tabular}{l | c| c}
\hline
Operation & Energy ($\rm pJ$) & Area ($\rm \mu m^2$)\\
\hline
16-bit Floating-point Add & 0.4 & 1360\\
16-bit Floating-point Mult & 1.1 & 1640\\
32-bit Floating-point Add & 0.9 & 4184\\
32-bit Floating-point Mult & 3.7 & 7700\\
\hline
8-bit Fixed-point Add & 0.03 & 36 \\
8-bit Fixed-point Mult & 0.2 & 282\\
32-bit Fixed-point Add & 0.1 & 137 \\
32-bit Fixed-point Mult & 3.1 & 3495\\
\hline
\end{tabular}
}
\end{center}
\end{table}

In particular, network quantization aims to project floating-point values onto a spaced grid, where the original floating-point values can be approximated by a set of discrete values. In this way, the compute-intensive floating-point operations can be replaced by power-efficient fixed-point or bitwise operations, which greatly reduces the computational cost of the networks.

Recently, many quantization methods~\cite{zhuang2018towards, Jung_2019_CVPR, Esser2020LEARNED} have been proposed and achieved promising results on some tasks such as image classification.
However, using aggressively low-bit quantized networks for more complex tasks such as object detection still remains a challenge. 
Developing quantized object detectors is a challenging task since a detector  not only performs object classification, but also needs to predict other rich information, such as the locations of bounding boxes for regression.
Some existing quantized object detection methods~\cite{jacob2018quantization,Zhuang_2020_CVPR} reduce the precision of detectors to 4 or 8 bits and achieve promising performance. However, when it comes to aggressively low bitwidth (\eg, 2-bit) quantization, directly quantizing the detector incurs a significant performance drop compared to their full-precision counterpart. Moreover, some layers (\eg, batch normalization, and skip connections) in the network still require floating-point arithmetic units for inference. This means both integer and floating-point arithmetic units are needed for inference.
As shown in Table~\ref{table:hardare_cost}, compared with fixed-point operations, floating-point operations consume much higher energy and area cost. 
Besides, data exchange between different types of arithmetic units may further hamper the energy efficiency of the network.

In this paper, we propose an Accurate Quantized object Detection (AQD) method to fully get rid of floating-point computation while maintaining performance. To this end, we propose to replace floating-point operations with fixed-point operations
in all kinds of layers, including the convolutional layers, normalization layers and skip connections. In this way, only integer arithmetic is required during inference, which significantly reduces the computational overheads. To reduce the performance drop from quantization while ensuring pure fixed-point operations, we further propose a new variant of batch normalization (BN) called multi-level BN. 
Our proposed method is based on the observation 
there is a large divergence of batch statistics across different feature pyramid levels, where batch statistics are computed using aggressively quantized activations. Therefore, using shared BN statistics in conventional detection frameworks \cite{tian2019fcos, lin2017focal} will result in highly poor estimates of statistical quantities.
To capture accurate batch statistics, multi-level BN privatizes batch normalization layers for each pyramid level of the head.

Our main contributions are summarized as follows:
\begin{itemize}\itemsep 0cm 
    \item
    We propose an Accurate Quantized object Detection (AQD) method to fully get rid of floating-point computation in each layer of the network, including convolutional layers, normalization layers and skip connections. As a result, only integer arithmetic is required during inference, which greatly improves the on-device efficiency to carry out inference.

    \item
	We highlight that the degraded performance of the quantized detectors is largely due to the inaccurate batch statistics in the network. We therefore propose multi-level batch normalization to capture accurate batch statistics of different levels of feature pyramid. 
	
	\item
	We evaluate the proposed methods on the COCO detection benchmark with multiple precisions. Experimental results show that our low bit AQD can achieve comparable or even better performance with its full-precision counterpart. 
\end{itemize}

\section{Related work}
	
	\noindent\textbf{Network quantization.} 
	Network quantization aims to reduce the model size and computational cost by
	representing the network weights and/or activations with low precision.
	Existing methods can be divided into two categories, namely, binary quantization~\cite{hubara2016binarized,rastegari2016xnor,bulat2018hierarchical,liu2018bi,liu2020reactnet} and fixed-point quantization~\cite{zhou2016dorefa, zhuang2018towards,zhang2018lq,Jung_2019_CVPR,Esser2020LEARNED}. Binary quantization converts the full-precision weights and activations to $\{+1, -1\}$, where the convolution operations are replaced with efficient bitwise operations and can achieve up to $32\times$ memory saving and $58\times$ speedup on CPUs~\cite{rastegari2016xnor,zhuang2019structured}. 
    To reduce the performance gap between the quantized model and the full-precision counterpart, fixed-point quantization methods~\cite{zhou2016dorefa,cai2017deep,zhuang2018towards,choi2018pact,zhang2018lq} represent weights and activations with higher bitwidths, showing impressive performance on the image classification task. 
    Besides,
    some logarithmic quantizers \cite{zhou2017incremental, miyashita2016convolutional, Li2020Additive} leverage 
    bit-shift operations
    to accelerate the computation. However, they impose constraints on the quantization algorithm. For example, the quantized activations need to be fixed-point values and the quantized weights are required to be powers-of-2 values, which might result in lower quantization performance.

    Apart from the algorithm design, the development of underlying implementation and acceleration libraries~\cite{cowan2018automating,zhang2019dabnn,alibaba2020mnn} are critical to enable highly-efficient execution on resource-constrained platforms.
    In particular, low-precision training methods~\cite{banner2018scalable,sun2019hybrid,wang2018training} quantize weights, activations and gradients to carefully designed data format for improved efficiency while preserving accuracy.
    To improve the inference efficiency, several methods~\cite{zhang2019dabnn,alibaba2020mnn} propose to design efficient bitwise operations on dedicated hardware devices, such as ARM, FPGA and ASIC.

	\noindent\textbf{Quantization on Object Detection}. Many researchers have studied quantization on object detection to speed up on-device inference and save storage.
	Wei~\etal\cite{Wei_2018_ECCV} utilize knowledge distillation and quantization to train very tiny CNNs for object detection. Zhuang~\etal\cite{Zhuang_2020_CVPR} point out the difficulty of propagating gradient and propose to train low-precision network with a full-precision auxiliary module. These works achieve promising quantization performance on object detection. However, they do not quantize all the layers (\eg, input and output layers, BN or skip connections), which limits the efficient deployment on resource-constrained platforms.
	Jacob~\etal\cite{jacob2018quantization} propose a quantization scheme using integer-only arithmetic and perform object detection on COCO dataset with 8-bit precision. Furthermore, when carrying out more aggressive quantization, Li~\etal\cite{Li_2019_CVPR} observe training instability during the quantized fine-tuning and propose three solutions.  
	However, these works impose extra constraints on both the network structure and quantization algorithm design, which limits them to obtain better performance (refer to Sec. \ref{sec:discuss}).
	In contrast, our quantization scheme is milder and more flexible in aspects of network structure and quantization algorithms which contributes to significant performance improvement of proposed AQD over several state-of-the-art quantized object detectors.

\section{Proposed method}
\subsection{Preliminary} 

\subsubsection{Integer-aware Quantized Representation
}
\label{sec:representation}

Full-precision (32-bit) floating-point is a general data type that is used to represent data in deep learning models. With network quantization, the continuous full-precision %
weights and activations are discretized to a limited number of quantized values. Specifically, given a specific bitwidth $b$, the total number of quantized values is $2^b$. To enable efficient integer arithmetic operations on the quantized values, it requires the quantization scheme to be a mapping of real values to integers.
Formally, for $b$-bit quantization of any full-precision value $x$ in a tensor $\bf{X}$,
its quantized version $\overline{x}$ 
can be formulated as:
\begin{equation}
\label{eq:represnet-linear}
\overline{x} = \eta \cdot \alpha,
\end{equation}
where $\alpha$ is a full-precision floating-point scale factor shared for the whole tensor $\bf{X}$, and $\eta \in {\mathbb{N}}$ implies the corresponding mapping value in the integer domain.

\subsubsection{Quantization function}
\label{sec:function}

In this work, we propose to quantize both weights and activations with learnable quantization intervals motivated by LSQ~\cite{Esser2020LEARNED}.
Without loss of generality, given a convolutional layer or fully-connected layer in a network, the weight $\mathbf{W}$ is convolved with the input activation $\mathbf{X}$, where $\mathbf{W}$ and $\mathbf{X}$ are real-valued tensors.
We use $x, w \in \mathbb{R}$ to denote the element of $\bX$ and $\bW$ respectively. 
Let $\nu_x$ and $\nu_w$ be the trainable quantization interval parameters that indicate the range of activations and weights to be quantized, which are shared for all elements in $\bX$ and $\bW$, respectively.

For a given $x$, we first constrain it to the range $[0, \nu_x]$, with values out of the range clipped into boundaries. We then linearly map values in the interval $[0, \nu_x]$ to the integer domain of $\{0,1,\hdots, 2^b-1\}$, where $b$ is the quantization bitwidth. At last, we restore the magnitude of the original $x$ by multiplying the corresponding scale factor. Formally, the quantization process can be formulated as follows:
\begin{equation}
\label{eq:mapping-fm}
\begin{aligned}
&\eta_x  = \lfloor {\rm clip} (\frac{x}{\nu_x}, 0, 1) \cdot (2^b -1) \rceil, \\
&  \overline{x} =  \eta_x \cdot \frac{\nu_x}{2^b-1},
\end{aligned}
\end{equation}
where $\lfloor \cdot \rceil$ returns the nearest integer of a given value, $\mathrm{clip}\left(x, x_\mathrm{low}, x_\mathrm{up}\right) = {\rm min}({\rm max}(x,x_\mathrm{low}),x_\mathrm{up})$, 
$\eta_x$ is the corresponding integer-domain mapping value of $x$, and $\frac{\nu_x}{2^b-1}$ is the magnitude-restore scale factor. 

For the given $w$ from weights, the quantization interval is defined as $[-\nu_w, \nu_w]$.
We compute the quantized weight $\overline{w}$ similar to $\overline{x}$ except that an additional transformation is applied 
, which can be formulated as:
\begin{equation}
\label{eq:mapping-wt}
\begin{aligned}
&\eta_w = \lfloor ({\rm clip} (\frac{w}{\nu_w}, -1, 1) + 1) /2 \cdot (2^b -1) \rceil, \\
    & \overline{w} = (\eta_w \cdot \frac{1}{2^b-1} \cdot 2 -1) \cdot \nu_w, 
\end{aligned}
\end{equation}
where $\eta_w$ is the corresponding integer-domain mapping value of $w$. 

During network training, the discretization operation by $\rm rounding$ function $\lfloor \cdot \rceil$ is non-differentiable. To avoid gradient vanishing issue, we employ the Straight-Through Estimator (STE) for back-propagation \cite{bengio2013estimating}.

\subsection{Floating-point Free Quantization}
\begin{figure}[!t] 
    \centering
	\includegraphics[width=0.95\linewidth]{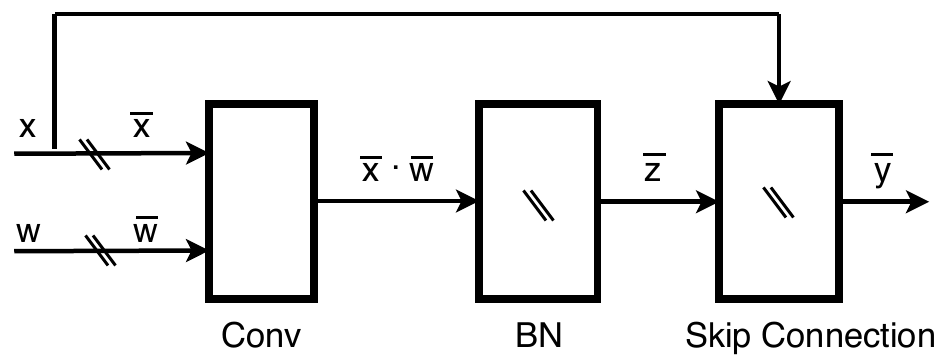}
	\caption{Illustration of a typical block in the detection network. Double slash lines indicate the positions where quantization applies. Note that in a fully-quantized network, $x$ can be represented in the format of Eq. (\ref{eq:represnet-linear}), which is the output $\overline{y}$ of the preceding block.} 
	\label{fig:quant-flow} 
\end{figure}

In this paper, we present an Accurate Quantized object Detection (AQD) method that fully gets rid of floating-point computation while still achieving promising performance. 
In a conventional residual block as shown in Figure \ref{fig:quant-flow}, we propose to substitute floating-point operations for fixed-point operations 
in all kinds of layers, including the convolutional layer, batch normalization layer and the skip connection. In this way, only integer arithmetic is required during network inference, which greatly reduces the computational overhead and memory footprint of the network. In the following subsections, we will introduce the details of the proposed method regarding fixed-point operations for each of these layers.

\begin{figure*}[t] 
	\centering
	
	\subfigure[3-th layer of classification head]{ 
		\label{fig:first_layer_cls_head} 
		\includegraphics[width=0.23\linewidth]{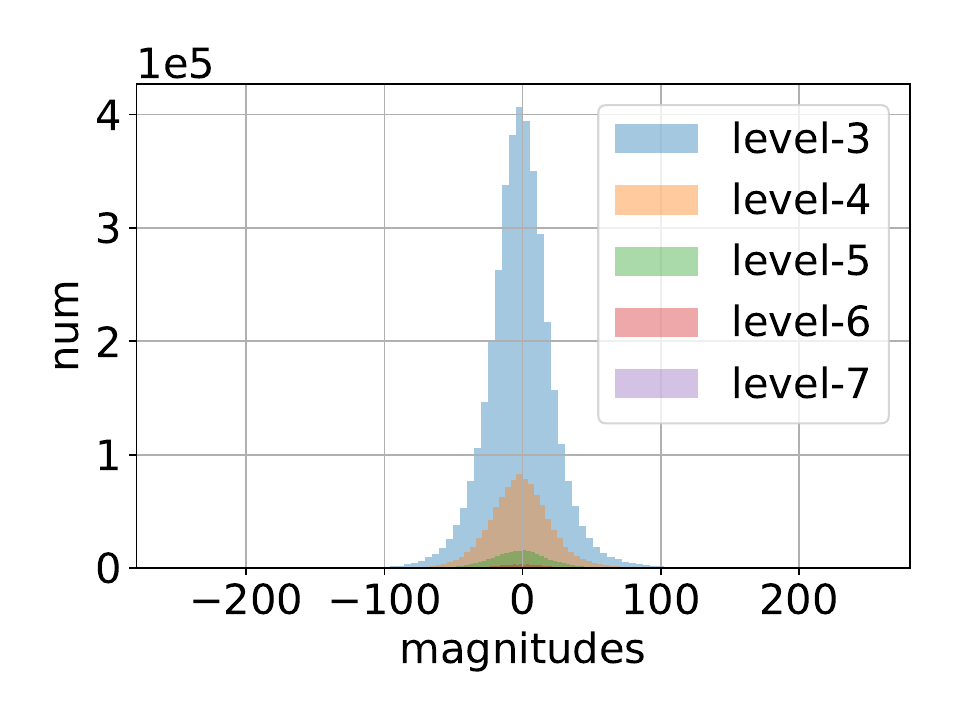}
		} 
	\subfigure[3-th layer of regression head]{ 
		\label{fig:first_layer_reg_head} 
		\includegraphics[width=0.23\linewidth]{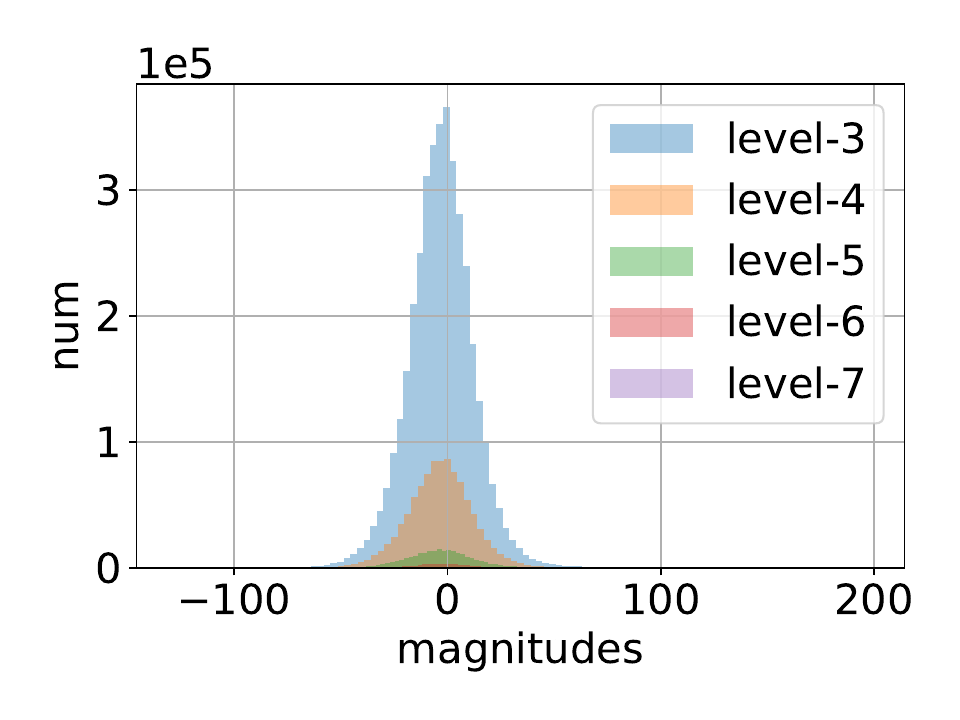}
		} 
	\subfigure[4-th layer of classification head]{ 
		\label{fig:second_layer_cls_head}  
		\includegraphics[width=0.23\linewidth]{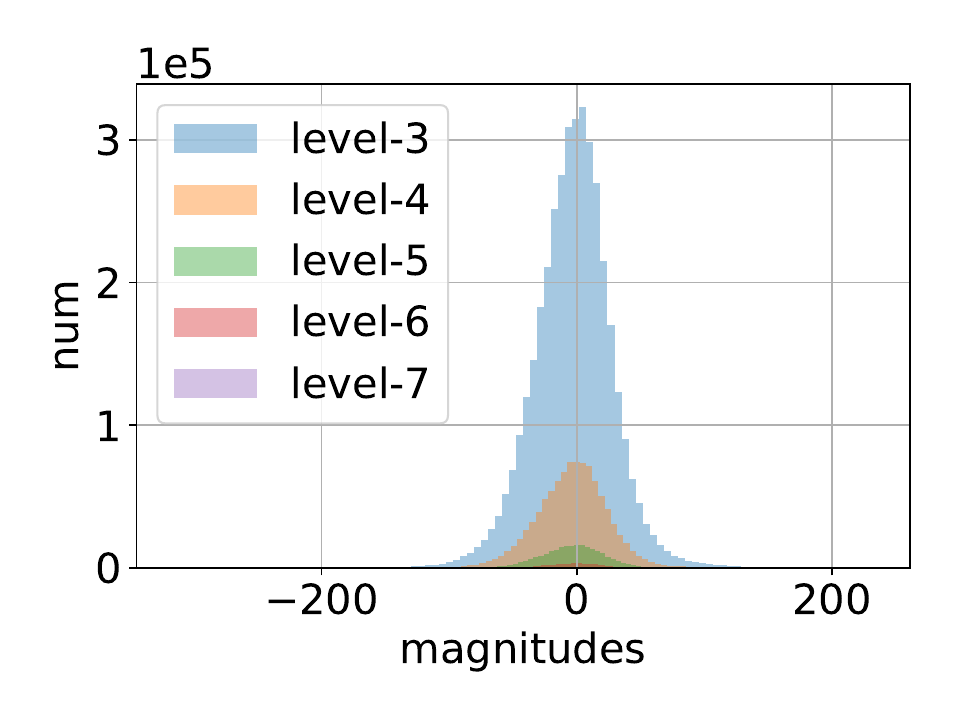}
		} 
	\subfigure[4-th layer of regression head]{ 
		\label{fig:second_layer_reg_head} 
		\includegraphics[width=0.23\linewidth]{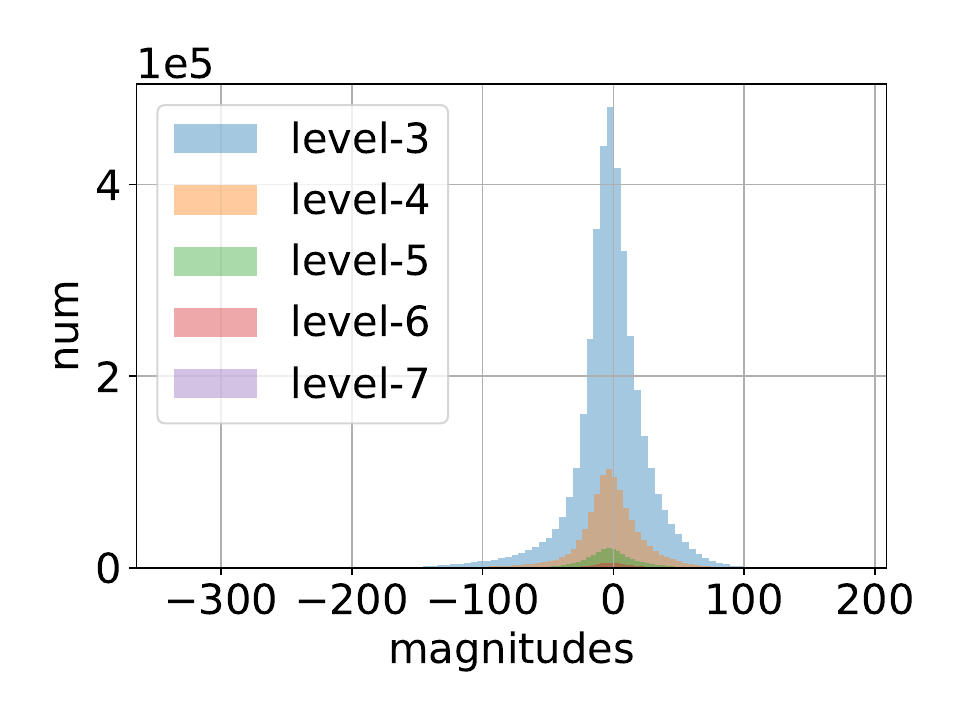}
		} 
	\caption{
	Distribution of input activations at the batch normalization layer in the detection heads of a 2-bit ResNet-18 FCOS detector. Level-x denotes that the predictions are made on the x-th pyramid level. Different levels of features show different batch of statistics.} 
	\label{fig:histograms_batch_statistics} 
\end{figure*}

\subsubsection{Convolutional and Fully-connected Layers}
\label{sec:conv}

As elaborated in Sec.~\ref{sec:representation},  we define a unified quantization representation that allows for integer-only arithmetic, and all tensors in the quantized network are required to follow the rule 
in Eq. (\ref{eq:represnet-linear}).
We can observe that the quantized activation $\overline{x}$ and weight $ \overline{w}$ are compatible with this representation format. 
Furthermore,
the output activation of a quantized convolutional or fully-connected layer can fit the format as well:

\begin{equation}
\label{eq:mapping-conv}
\begin{aligned}
\overline{x} \cdot \overline{w} &= (\eta_x \cdot \frac{\nu_x}{2^b-1}) \cdot ((\eta_w \cdot \frac{1}{2^b-1} \cdot 2 -1) \cdot \nu_w), \\
 & = (\eta_x \cdot (2 \cdot \eta_w - 2^b + 1)) \cdot \frac{\nu_x \cdot \nu_w}{(2^b-1)^2}, \\
 & = \eta_{\rm{conv}} \cdot \alpha_{\rm{conv}}.
\end{aligned}
\end{equation}
Benefit from the data representation in Eq. (\ref{eq:represnet-linear}), the scale factors $\nu_x$ and $\nu_w$ can be handled independently from $\eta_x$ and $\eta_w$. 
Being shared for the whole activations and fixed in inference,
the floating-point scale factor $\alpha_{\rm conv}$
can be passed to the proceeding layer directly. In this case, only fixed-point operations exist in the quantized convolutional or fully-connected layer, which can greatly reduce the computational cost.

\subsubsection{Normalization Layers}
\label{sec:normalization}

\begin{figure}[!htb] 
    \centering
	\includegraphics[width=0.95\linewidth]{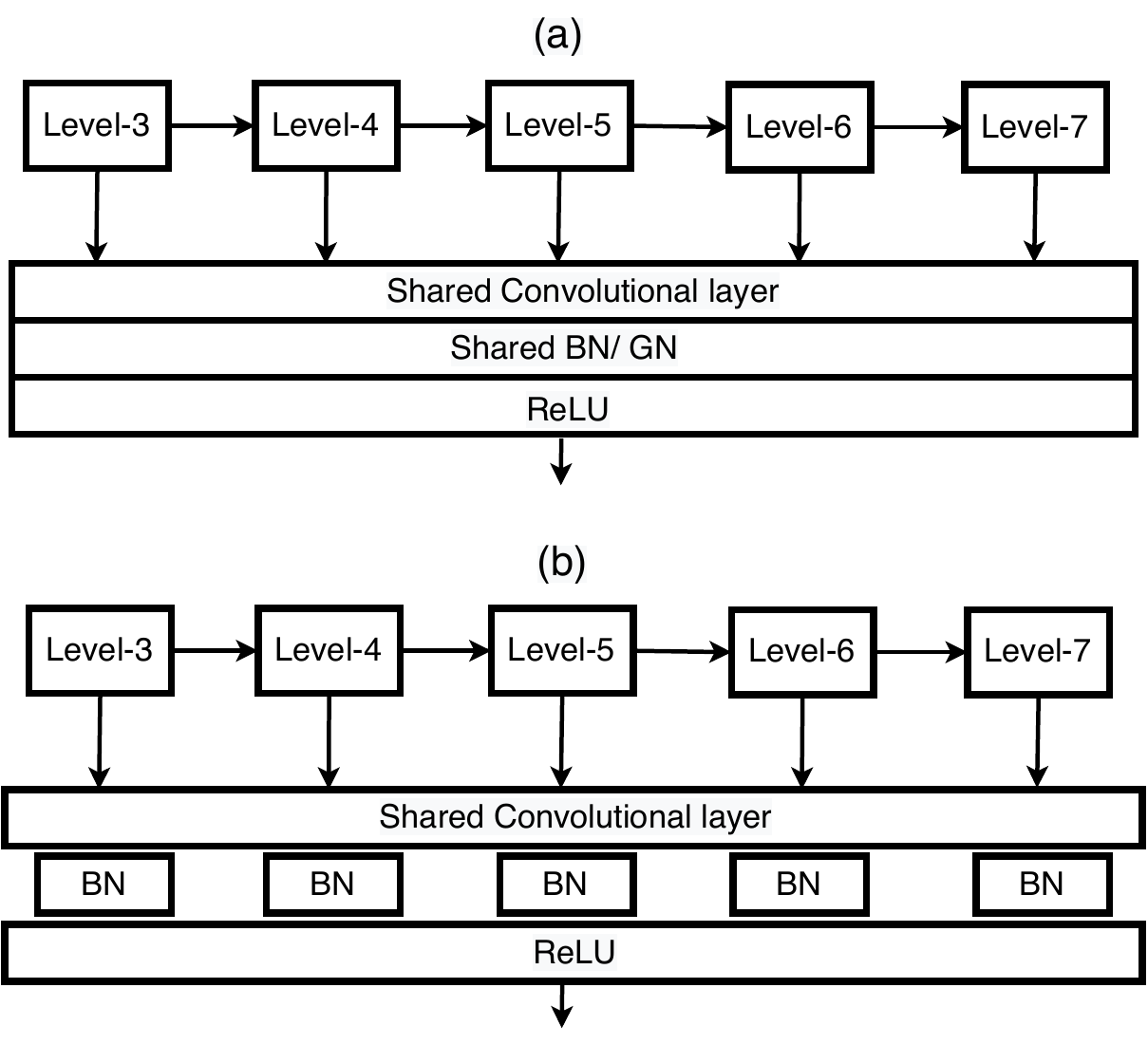}
	\caption{Illustration of the proposed quantized detection heads design. (a) Conventional detection frameworks with shared group normalization or batch normalization \cite{tian2019fcos,lin2017focal}; (b) The proposed multi-level BN that privatizes batch normalization layers for different heads.} 
	\label{fig:multi-level-BN} 
\end{figure}

Both batch normalization (BN)~\cite{ioffe2015batch}
and group normalization (GN)~\cite{wu2018group} are widely used in classical object detection networks \cite{lin2017focal, tian2019fcos, law2018cornernet}. We investigate the differences of these two normalization operations and analyze their impact on network quantization and inference. 

In particular, BN is employed mainly in the backbone module for feature extraction while GN is preferred in the detection head, which is generally shared to handle extracted feature of different levels in the feature pyramid network (FPN) \cite{lin2017feature}. Both BN and GN normalize the input activations using mean and variance with an optional affine transformation. However, these two normalization strategies differ in two aspects.
On the one hand, a BN layer calculates the statistics of the whole mini-batch while those of GN only reply on the individual input tensor in the mini-batch. Thus, BN is more sensitive to the training batch size, and might suffer from inaccurate statistics when the batch size is small. In contrast, GN behaves more stable to the training batch size.
On the other hand, BN keeps track of exponential moving average mean $\mu$ and variance $\sigma$, and updates them using the minibatch statistics at each forward step during training, while $\mu$ and $\sigma$ are \textbf{fixed} during inference. 
In contrast, GN \textbf{re-computes} the statistics information in each forward step during both training and inference procedures. As a result, GN unavoidably incurs floating-point computation during inference. Moreover, compared with BN, which can be potentially fused into the corresponding convolutional layer with no extra computational cost, GN imposes more execution burden during inference.

Therefore, to allow for integer-only arithmetic, we suggest to use batch normalization to replace group normalization in quantized detection networks.
However, this replacement might cause training instability since batch statistics are computed using aggressively quantized activations, especially a small training batch size is commonly used. %
Fortunately, when equipped with the synchronized version of batch normalization (Sync-BN), the inaccurate statistics issue of batch normalization can be partially addressed \cite{peng2018megdet}.
However, another issue may occur when FPN \cite{lin2017feature} is employed in the detectors.
We can observe a large divergence of batch statistics between different feature pyramid levels, as shown in Figure~\ref{fig:histograms_batch_statistics}.
Therefore, using shared BN statistics across prediction heads may lead to highly degraded estimates of statistical quantities, which will cause a significant performance drop. 
To solve this issue, we propose 
multi-level batch normalization (multi-level BN) to attain better quantization performance. Specifically, multi-level BN privatizes batch normalization layers for the shared convolutional layers in different detection heads, as illustrated in Figure \ref{fig:multi-level-BN}. 
The proposed multi-level BN can capture individual batch statistics of the corresponding feature level. Furthermore, only negligible parameters (less than 1.1\% of the model size) are introduced. Besides, it should be noted that only one batch normalization is activated at a time though multiple batch normalizations are allocated (since extracted features from different levels are computed sequentially). Thus, the proposed multi-level BN will not increase the computational cost compared with the traditional shared batch normalization (or group normalization).

To enable integer-only computation, we now design the quantization strategy for the batch normalization layer to remove the floating-point operations.
We assume the previous layer is a quantized convolutional or fully-connected layer, whose output is $\overline{x} \cdot \overline{w}$ as explained in Eq. (\ref{eq:mapping-conv}).
Batch normalization layer then takes it as input and obtains the normalized output $z$.
Before quantizing the batch normalization layer, the normalized value $z$ can be computed by  

\begin{equation}
\begin{aligned}
\label{eq:bn-3}
z &= \frac{\eta_{\rm{conv}} \cdot \alpha_{\rm{conv}} - \mu}{\sqrt{\sigma^2 + \epsilon}} \cdot \gamma + \beta, \\
 &= (\eta_{\rm{conv}} + \frac{\beta \cdot \frac{\sqrt{\sigma^2 + \epsilon}}{\gamma} - \mu}{\alpha_{\rm{conv}}}) \cdot \frac{\alpha_{\rm{conv}} \cdot \gamma}{\sqrt{\sigma^2 + \epsilon}}, \\
 &= (\eta_{\rm{conv}} + s) \cdot \alpha_z,
\end{aligned}
\end{equation}
where $\gamma$, $\beta$ are the affine transformation parameters, $\epsilon$ is a constant to stabilize the computation.
Then we are able to obtain the quantized batch normalization layer by quantizing $s$:
\begin{equation}
\overline{z} = (\eta_{\rm{conv}} + \lfloor s \rceil) \cdot \alpha_z = \eta_z \cdot \alpha_z .
\end{equation}
Obviously, for quantized batch normalization, the main computation is the $\rm add$ operation between $\eta_{\rm{conv}}$ and $\lfloor s \rceil$ and the floating-point factor $\alpha_z$ can be passed to next layer.

It is worth noting that, compared with the output of the quantized convolutional or fully-connected layer, the quantized output $\overline{z}$ of the batch normalization layer also meets 
the integer-aware representation in Eq. (\ref{eq:represnet-linear}), but with the channel-wise scale factor $\alpha_z$.

\subsubsection{Skip Connections}
\label{sec:skip_connection}
 
Skip connections are commonly applied in the detectors with the ResNet \cite{he2016deep} backbone and the FPN module. In this subsection, we elaborate how to enable integer-only computation of skip connections (\ie, $\rm add$ operation between two tensors). Diving into the structure of the object detection network, we can learn that skip connections are employed mainly in three occasions: 1) Fusion of the identity mapping $x$ and batch normalization output $\overline{z}$ in a residual block as illustrated in Figure \ref{fig:quant-flow}. 2) Fusion of the batch normalization output of the downsampling branch and the main branch in a residual block. 3) Fusion of the up-scaled feature map and the current level feature map in FPN. 
For either of these occasions, given two input quantized values $x_1$ and $x_2$, the output  $y$ of skip connection and its quantized value $\overline{y}$ are computed as follows: %
 \begin{equation}
 \label{eq:skip-1}
 \begin{aligned}
 y &= x_1 + x_2 = \eta_{1} \cdot \alpha_{1} + \eta_{2} \cdot \alpha_{2}, \\
 \overline{y} &= \left\{
\begin{array}{ll}
(\eta_{1} + \eta_{2} \cdot F(\alpha_{2}, \alpha_{1}) ) \cdot \alpha_{1}  & {\rm{if}} \;  \alpha_{2} \geq \alpha_{1} \\
(\eta_{1} \cdot F(\alpha_1, \alpha_2) + \eta_{2} ) \cdot \alpha_2 & \rm{otherwise}
\end{array}
\right.,
 \end{aligned}
 \end{equation}
 where function $F(\cdot)$ is defined as
 
\begin{equation}
\label{eq:skip-3}
F(m, n) = \frac{c}{2^d}, {{\rm where}}~c, d = \argmin_{c \in \mathbb{N}, d \in \mathbb{N_+}} {| \frac{m}{n}  - \frac{c}{2^d}|}.
 \end{equation}

Function $F(\cdot)$ actually is used for searching an approximation of a given fraction $\frac{m}{n}$ (where $m \geq n$). 
In particular, the numerator $c$ is constrained to be an integer and the denominator $2^d$ is constrained to be powers-of-two to allow for bit-shift ($d$ is a positive integer). Function $F(\cdot)$ finds the best combination of $c$ and $d$ to approximate $\frac{m}{n}$. There is limited choice for $d$, for example $d \in \{0, 1, 2, \hdots, 31\}$ if $\eta_1$ and $\eta_2$ are represented in $32$-bit integer.
Therefore, function $F(\cdot)$ can be quickly solved, and Eq. (\ref{eq:skip-1}) can be implemented with fixed-point operations only (the scale factors $\alpha_1$ and $\alpha_2$ do not participate in the computation and are passed to the next layer directly). It is worth mentioning that Eq. (\ref{eq:skip-1}) is conducted in a channel-wise manner since the scale factor is applied on each channel in the batch normalization layer. 

 \subsubsection{Other Layers}
 
 Layers such as max-pooling and nearest interpolation do not introduce floating-point computation. Besides, they do not change the scale factors of the layer input. Thus, there is no special modifications for these layers.

\subsection{Discussions}
\label{sec:discuss}
Related to our work, FQN \cite{Li_2019_CVPR} also targets on floating-point-free arithmetic operations during inference. 
Compared with FQN, our method imposes milder constraints for both the network structure and quantization algorithm design, to get rid of floating-point computation.

On the one hand, instead of quantizing the batch normalization in standalone during training, FQN employs batch normalization folding (BN folding), which fuses the batch normalization layer into the preceding convolutional layer to simulate the quantization effect.
However, as indicated in FQN, training instability is observed for the batch normalization parameters and batch statistics after BN folding is leveraged since considerable quantization noise is introduced especially for the extremely low bitwidth. To address this issue, FQN proposes
\emph{freezed batch normalization} in which batch statistics are fixed during quantized finetuning.
Instead, we propose to employ the standard batch normalization with statistics and parameters updated, but with the multi-level design.

On the other hand, FQN imposes constraints on the quantization algorithm. In particular, they require the scale factors of the two quantized input activations in the skip connection to satisfy

\begin{equation}
\label{eq:discuss-1}
\frac{\alpha_1}{\alpha_2} = 2^d,
 \end{equation}
 where $d$ is an integer (different with the scope in Eq. (\ref{eq:skip-3}), $d$ here can be either positive or negative). This constraint is equivalent to Eq. (\ref{eq:skip-3}) by fixing $c$ to be constant 1. To meet such a condition, dedicated design of the quantization algorithm for the preceding
 convolutional layers is required due to the scale factors.
In contrast, our method does not make such an assumption. Actually, the proposed method allows to quantize different layers independently.
With the proposed simple yet mild design, our method shows superior performance over FQN as demonstrated in Sec. \ref{sec:exp}.

\section{Experiments}
\label{sec:exp}

\begin{table*}[t]
	\begin{center}
	\caption{Performance comparisons on the COCO validation set based on RetinaNet. 
	}
	\label{table:results_retinanet_coco}
	\small 
	 \scalebox{0.90}
	{
	\begin{tabular}{c|c|ccc|ccc}
	\hline
	Backbone & Method & AP & AP$_{50}$ & AP$_{75}$ & AP$_S$ & AP$_M$ & AP$_L$\\
	\hline
	\multirow{9}{*}{\tabincell{c}{ResNet-18}} 
	& Full-precision & 32.3 & 50.9 & 34.2 & 18.9 & 35.6 & 42.5\\
	\cdashline{2-8}
	& FQN~\cite{Li_2019_CVPR} (4-bit) & 28.6 & 46.9 & 29.9 & 14.9 & 31.2 & 38.7 \\
	& Auxi~\cite{Zhuang_2020_CVPR} (4-bit) & 31.9 & 50.4 & 33.7 & 16.5 & 34.6 & 42.3 \\
	\cdashline{2-8}
	& AQD* (4-bit) & 34.1 & 53.4 & 36.4 & 19.8 & 36.4 & 44.7 \\
	& AQD  (4-bit) & 34.1 & 53.1 & 36.3 & 19.4 & 36.4 & 45.0 \\
	\cdashline{2-8}
	& AQD* (3-bit) & 33.5 & 52.5 & 35.6 & 17.8 & 35.9 & 44.9 \\
	& AQD (3-bit) &  33.4 & 52.8 & 35.7 & 17.9 & 36.4 & 43.9 \\
	\cdashline{2-8}
	& AQD* (2-bit) & 31.0 & 49.5 & 32.7& 17.0 & 33.1& 41.4 \\
	& AQD (2-bit) & 30.8 & 50.0 & 32.3 & 16.5 & 33.1 & 41.5  \\
    \hline 
	
	\multirow{9}{*}{\tabincell{c}{ResNet-34}} 
	& Full-precision & 36.3 & 56.2 & 39.1 & 22.4 & 39.8 & 46.9\\
	\cdashline{2-8}
	& FQN~\cite{Li_2019_CVPR} (4-bit) & 31.3 & 50.4& 33.3 & 16.1 & 34.4 & 41.6 \\
	& Auxi~\cite{Zhuang_2020_CVPR} (4-bit) & 34.7 & 53.7 & 36.9 & 19.3 & 38.0 & 45.9 \\
	\cdashline{2-8}
	& AQD* (4-bit) & 37.1 & 56.8 & 40.0 & 21.8 & 40.3 & 48.1 \\
	& AQD  (4-bit) & 37.1 & 56.8 & 39.8 & 21.9 & 40.0 & 48.0 \\
	\cdashline{2-8}
	& AQD* (3-bit) & 36.5 & 56.3 & 38.9 & 21.2 & 39.4 & 48.2 \\
	& AQD (3-bit) &  36.5 & 56.3 & 38.8 & 21.4 & 39.5 & 47.7 \\
	\cdashline{2-8}
	& AQD* (2-bit) & 34.3 & 53.8 & 36.4 & 19.6 & 37.0 & 45.3  \\
	& AQD (2-bit) & 33.8 & 54.0 & 36.1 & 19.4 & 36.8 & 44.8   \\
	\hline
	
	\multirow{9}{*}{\tabincell{c}{ResNet-50}} 
	& Full-precision & 37.8 & 58.0 & 40.8 & 23.8 & 41.6 & 48.9\\
	\cdashline{2-8}
	& FQN~\cite{Li_2019_CVPR} (4-bit) & 32.5 & 51.5& 34.7 & 17.3 & 35.6 & 42.6 \\
	& Auxi~\cite{Zhuang_2020_CVPR} (4-bit) & 36.1 & 55.8 & 38.9 & 21.2 & 39.9 & 46.3 \\
	\cdashline{2-8}
    & AQD* (4-bit) & 38.1 & 58.5 & 41.3 & 23.9 & 41.8 & 48.7 \\
    & AQD  (4-bit) & 38.1 & 58.1 & 40.7 & 22.5 & 41.6 & 49.8 \\
	\cdashline{2-8}
	& AQD* (3-bit) & 37.2 & 57.4 & 39.5 & 23.0 & 40.8 & 47.8 \\
	& AQD (3-bit) &  36.9 & 57.1 & 39.5 & 22.0 & 40.6 & 47.9 \\
	\cdashline{2-8}
	& AQD* (2-bit) & 35.0 & 55.0 & 37.2 & 20.6 & 38.4 & 45.5 \\
	& AQD (2-bit) &  34.8 & 55.4 & 36.9 & 20.3 & 37.9 & 45.6 \\
	\hline
    \end{tabular}}
	\end{center}
	\vspace{-4mm}
\end{table*}
 
\begin{table*}[htp]
\begin{center}
\caption{Effect of the multi-level batch normalization. We evaluate performance of both full-precision (FP) and 2-bit quantized models based on FCOS on the COCO validation set. 
}
\label{table:effect_bn}
\scalebox{0.80}{
\begin{tabular}{c|c|c|cc|ccc|cccc}
\hline
Backbone & Normalization & Precision & Shared & Fixed-point-only & AP & AP$_{50}$ & AP$_{75}$ & AP$_S$ & AP$_M$ & AP$_L$\\
\hline
\multirow{6}{*}{\tabincell{c}{ResNet-18}} 
	& BatchNorm &  \multirow{3}{*}{\tabincell{c}{FP}}  & \checkmark  &  & 29.5 & 46.6 & 31.7 & 19.0 & 32.8 & 35.8 \\
	& GroupNorm &  & \checkmark &  & 34.0 & 51.7 & 36.3 & 19.7 & 36.6 & 44.0 \\
	& Multi-level BatchNorm &  &  &  & 33.9 & 51.2& 36.4& 19.3& 36.2& 44.0 \\
\cdashline{2-11}
	& BatchNorm & \multirow{3}{*}{\tabincell{c}{2-bit}}  & \checkmark  & \checkmark  & 26.4 & 43.6 & 28.2 & 14.3 & 28.7 & 34.6 \\
	& GroupNorm &  & \checkmark &  & 29.4 & 47.2 & 31.7 & 15.4 & 31.6 & 38.6 \\
	& Multi-level BatchNorm &  &  & \checkmark & 31.8 & 49.3 & 34.2 & 17.3 & 33.5 & 42.3 \\
	
\hline

\multirow{6}{*}{\tabincell{c}{ResNet-50}} 
	& BatchNorm & \multirow{3}{*}{\tabincell{c}{FP}}  & \checkmark &  & 35.9 & 53.9 & 39.0 & 21.9 & 39.2 & 45.8 \\
	& GroupNorm &  & \checkmark &  & 38.7 & 57.6 & 41.4 & 22.8 & 42.3 & 50.2 \\
	& Multi-level BatchNorm & & &  & 38.9 & 57.4 & 42.1 & 23.6 & 42.0 & 50.3\\
\cdashline{2-11}
	& BatchNorm & \multirow{3}{*}{\tabincell{c}{2-bit}}  & \checkmark &\checkmark & 30.3 & 49.6 & 32.6 & 17.3 & 32.9 & 39.0 \\
	& GroupNorm &  & \checkmark &  & 33.4 & 52.2 & 35.8 & 18.4 & 37.2 & 42.5 \\
	& Multi-level BatchNorm  &  & & \checkmark  & 35.4 & 54.1 & 38.2 & 19.5 & 38.0 & 46.2 \\
\hline
\end{tabular}
} %
\end{center}
	\vspace{-2mm}
\end{table*}

\begin{table*}[t]
	\begin{center}
	\caption{Performance comparisons on the COCO validation set based on FCOS.}
	\label{table:results_fcos_coco}
	\scalebox{0.90}
	{
	\small 
	\begin{tabular}{c|c|ccc|ccc}
	\hline
	Backbone & Model & AP & AP$_{50}$ & AP$_{75}$ & AP$_S$ & AP$_M$ & AP$_L$\\
	\hline
	\multirow{8}{*}{\tabincell{c}{ResNet-18}} 
	& Full-precision & 33.9 & 51.2& 36.4& 19.3& 36.2& 44.0\\
	\cdashline{2-8}
	& Group-Net~\cite{zhuang2019structured} (4 bases) &28.9  &45.3  &31.2  &15.4  &30.5  &38.1  \\
	\cdashline{2-8}
	& AQD* (4-bit) & 34.9 & 52.1 & 37.3 & 19.9 & 36.5 & 45.6 \\
	& AQD  (4-bit) & 34.1 & 51.5 & 36.5 & 18.3 & 36.2 & 45.1 \\
	\cdashline{2-8}
	& AQD* (3-bit) & 34.3 & 51.4 & 36.7 & 19.4 & 36.0 & 45.1 \\
	& AQD (3-bit) & 33.6 & 51.1 & 36.1 & 18.5 & 35.0 & 44.8 \\
	\cdashline{2-8}
	& AQD* (2-bit) & 32.2 & 49.0 & 34.1 & 17.6 & 33.7 & 42.7 \\
	& AQD (2-bit) & 31.8 & 49.3 & 34.2 & 17.3 & 33.5 & 42.3  \\
	\hline
	
    \multirow{8}{*}{\tabincell{c}{ResNet-34}} &
    Full-precision & 38.0 & 55.9 & 41.0 & 23.0 & 40.3 & 49.4\\
    \cdashline{2-8}
    & Group-Net~\cite{zhuang2019structured} (4 bases) &31.5  &47.6  &33.8  &16.9  &32.3  &40.1  \\
    \cdashline{2-8}
	& AQD* (4-bit) & 38.3 & 56.2 & 41.3 & 21.8 & 40.5 & 49.8 \\
	& AQD  (4-bit) & 37.6 & 55.5 & 40.6 & 20.8 & 40.0 & 49.3 \\
	\cdashline{2-8}
	& AQD* (3-bit) & 37.8 & 55.8 & 40.7 & 22.2 & 40.4 & 49.8 \\
	& AQD (3-bit) &  37.2 & 55.2 & 40.2 & 20.6 & 39.5 & 48.8 \\
	\cdashline{2-8}
	& AQD* (2-bit) & 35.7 & 53.3 & 38.3 & 20.4 & 37.8 & 47.2 \\
	& AQD (2-bit) &  35.0 & 53.4 & 37.5 & 18.9 & 37.3 & 47.1  \\
	\hline
	
	\multirow{8}{*}{\tabincell{c}{ResNet-50}} 
	& Full-precision & 38.9 & 57.4 & 42.1 & 23.6 & 42.0 & 50.3\\
	\cdashline{2-8}
	& Group-Net~\cite{zhuang2019structured} (4 bases) & 32.7 &  49.0  & 35.5  & 17.8  &  33.6  & 41.4 \\
    \cdashline{2-8}
	& AQD* (4-bit) & 38.8 & 57.1 & 41.5 & 23.2 & 41.5 & 50.1 \\
	& AQD  (4-bit) & 38.0 & 56.5 & 40.7 & 21.9 & 41.0 & 49.2 \\
	\cdashline{2-8}
	& AQD* (3-bit) & 38.5 & 57.1 & 41.9 & 23.3 & 41.6 & 49.7 \\
	& AQD (3-bit) &  37.5 & 56.2 & 40.2 & 21.6 & 40.6 & 48.6 \\
	\cdashline{2-8}
    & AQD* (2-bit) & 36.0 & 53.8 & 38.8 & 20.0 & 38.6 & 47.0 \\
	& AQD (2-bit)  & 35.4 & 54.1 & 38.2 & 19.5 & 38.0 & 46.2 \\
	\hline
    \end{tabular}}
    \end{center}
	\vspace{-4mm}
\end{table*}

\begin{table*}[htp]
\begin{center}
\caption{Effect of quantization on different components. We quantize the FCOS detector to 2-bit and evaluate the performance on the COCO validation set based on AQD*.}
\label{table:different_components}
\scalebox{0.81}{
\begin{tabular}{c|c|ccc|ccc}
\hline
Backbone & Model & AP & AP$_{50}$ & AP$_{75}$ & AP$_S$ & AP$_M$ & AP$_L$\\
\hline
\multirow{4}{*}{\tabincell{c}{ResNet-18}} & Full-precision~\cite{tian2019fcos} & 33.9 & 51.2& 36.4& 19.3& 36.2& 44.0\\
	& Backbone & 33.8 & 50.6 & 36.1 & 18.6 & 35.4 & 45.4\\
	& Backbone + Feature Pyramid & 33.2 & 49.9 & 35.4 & 18.9 & 34.6 & 44.0\\ 
	& Backbone + Feature Pyramid + Heads & 32.2 & 49.0 & 34.1 & 17.6 & 33.7 & 42.7 \\
\hline
\end{tabular}
}	
\end{center}
	\vspace{-3mm}
\end{table*}

We evaluate our proposed method on the COCO detection benchmarks dataset~\cite{lin2014microsoft}. COCO detection benchmark is a large-scale benchmark dataset for object detection, which is widely used to evaluate the performance of detectors. Following~\cite{lin2017feature, zhuang2019structured}, we use the COCO \textit{trainval35k} split (115K images) for training and \textit{minival} split (5K images) for validation. 

\noindent\textbf{Comparison methods.}
To investigate the effectiveness of the proposed method, we compare with several state-of-the-art quantized object detection methods, including Auxi \cite{Zhuang_2020_CVPR}, Group-Net \cite{zhuang2019structured} and FQN \cite{Li_2019_CVPR}. We also define the following methods for study:
\textbf{AQD*}: Following~\cite{zhuang2019structured, Zhuang_2020_CVPR} we quantize all the convolutional layers, except the input layer in the backbone and the output layers in the detection heads, which acts as a strong baseline in our work. \textbf{AQD}: We quantize all network layers, including the input and output layers, batch normalization and skip connection layers, which is our complete method. Input and output layers are quantized to 8-bit in all settings.

\noindent\textbf{Implementation details.}
We implement the proposed method based on Facebook Detectron2~\cite{wu2019detectron2}. We apply the proposed AQD on two classical one-stage object detectors, namely, RetinaNet~\cite{lin2017focal} and FCOS~\cite{tian2019fcos}. 
We use ResNet~\cite{he2016deep} for the backbone module. %
To stabilize the optimization, we add batch normalization layers and ReLU non-linearities after the convolutional layers by default. 
For data pre-processing, we follow the strategy in ~\cite{Li_2019_CVPR, Zhuang_2020_CVPR} to resize images with a shorter edge to 800 pixels in the training and validation set.
Besides, images are augmented by random horizontal flipping during training. We do not perform any augmentations during evaluation. Total 90K iterations are trained with a mini-batch size of 16. We use SGD optimizer with a momentum of 0.9 for optimization.
The learning rate is initialized to 0.01, and divided by 10 at iterations 60K and 80K, respectively. We set the weight decay to 0.0001. More detailed settings on the other hyper-parameters can be found in \cite{lin2017focal,tian2019fcos}. Training hyper-parameters are the same for the quantized network and the full-precision counterpart, except for the initialization strategy. Specifically, to train the full-precision models, backbones are pre-trained on the ImageNet \cite{russakovsky2015imagenet} classification task. Parameters from other parts are randomly initialized. Whereas, for training the quantized detector, the whole network is initialized by the full-precision counterpart. 

\subsection{Comparison with State-of-the-art Methods}
	
We compare the proposed method with several state-of-the-arts and report the results in Tables~\ref{table:results_retinanet_coco} and \ref{table:results_fcos_coco}. 
Quantization performance of 4/3/2-bit networks are listed in the tables. Full-precision performance is also provided for comparison. 
In particular, Group-Net~\cite{zhuang2019structured} employs 4 binary bases, which corresponds to 2-bit fixed-point quantization.

Based on the results from Tables \ref{table:results_retinanet_coco} and \ref{table:results_fcos_coco}, we have the following observations. Firstly, our AQD consistently outperforms the compared baselines on different detection frameworks and backbones. For example, our 4-bit RetinaNet detector with ResNet-18 backbone outperforms FQN~\cite{Li_2019_CVPR} and Auxi~\cite{Zhuang_2020_CVPR} by 5.5\% and 2.2\% on AP, respectively. %
Besides, our 2-bit FCOS detector obtains 2.9\% and 2.7\% AP improvement over the Group-Net with ResNet-18 and ResNet-50 backbones, separately.
Moreover, our 4-bit quantized detectors can even outperform the corresponding full-precision models in some cases. Specifically, on a 4-bit RetinaNet detector, our AQD surpasses the full-precision model by 1.8\% and 0.8\% on AP with ResNet-18 and ResNet-34 backbones, respectively. Furthermore, when performing 3-bit quantization, our AQD achieves near lossless performance compared with the full-precision counterpart. For example, on a 3-bit RetinaNet detector with ResNet-50 backbone, our AQD only leads to 0.9\% performance degradation on AP. Lastly, when conducting aggressive 2-bit quantization, our AQD still achieves promising performance. For example, our fully-quantized 2-bit RetinaNet detector with ResNet-50 backbone only incurs 3.0\% AP loss compared with its full-precision baseline, but with considerable computation saving. These results justify the superior performance of our proposed AQD.

\subsection{Ablation Study}

\noindent\textbf{Effect of Multi-level Batch Normalization.}
\label{sec:effect_multi_level_bn}
To study the effect of multi-level BN, we quantize the FCOS detector with multi-level BN, shared GN (or BN) and report the results in  Table~\ref{table:effect_bn}. Here, the detector with shared GN (or BN) indicates that normalization layers in the detection heads are shared across different pyramid levels, which is the default setting in current prevalent detectors. In practical, synchronized version of batch normalization (Sync-BN) is leveraged for all batch normalization  layers. From the results, we have several observations. Firstly, the detector with the multi-level Sync-BN outperforms the one using the shared Sync-BN consistently by a large margin for both full-precision and quantized models with different backbones. %
Secondly, the detector with multi-level BN obtains comparable performance with the one with GN on full-precision models, and 2\% or more AP improvement %
on the 2-bit quantization,
which justifies the multi-level design can effectively solve the training instability during quantized fine-tuning.
Thirdly, compared with the group normalization, our multi-level BN performs much better on quantized models while enabling computation to be carried out using integer-only arithmetic, which is more hardware friendly.
	
\noindent\textbf{Effect of Quantization on Different Components.}
We further study the effect of quantizing different components in object detection models. The results are shown in Table~\ref{table:different_components}. We observe that quantizing the backbone or the feature pyramid only leads to a small performance drop. Nevertheless, quantizing the detection heads and the feature pyramid will cause significant performance degradation (\ie, 1.7\% in AP). These results show that the detection head modules other than the backbone are sensitive to quantization, which provides a direction to improve the performance of the quantized network.

\noindent\textbf{AQD vs. AQD*.}
We further study the influence of fully-quantizing a model compared to the one with only convolutional layers quantized.
From the results in Tables \ref{table:results_retinanet_coco} and \ref{table:results_fcos_coco}, we observe that AQD has a limited performance drop compared with AQD*. For example, the degradation is less than 0.3\% on RetinaNet with ResNet-18 and ResNet-50 backbones of different quantization bitwidths.
Besides, the performance gap on FCOS is relatively larger than that on RetinaNet. It can be attributed that FCOS is a fully convolutional pixel prediction framework relying heavily on the pixel-level feature quality, which might be more sensitive to the extreme low precision quantization.

\section{Conclusion}
In this paper, we have proposed an accurate quantized object detection framework with fully integer-arithmetic operations.
Specifically, we have proposed efficient integer-only operations for BN layers and skip connections. Moreover, we have proposed multi-level BN to accurately calculate batch statistics for each pyramid level. 
To evaluate the performance of the proposed methods, we have applied our AQD on two classical one-stage detectors. Experimental results have justified that our quantized 3-bit detector achieves comparable performance compared with the full-precision counterpart. More importantly, our 4-bit detector can even outperform the full-precision counterpart in some cases, which is of great practical value.

\section*{Acknowledgements}
\small 
MT was in part supported by Key-Area Research and Development Program of Guangdong Province 2018B010107001, and Program for Guangdong Introducing Innovative and Enterpreneurial Teams 2017ZT07X183, and Fundamental Research
Funds for the Central Universities D2191240. 
CS and his employer received no financial support for the research, authorship, and/or publication of this article.

{\small
\bibliographystyle{ieee_fullname}
\bibliography{reference}
}

\end{document}


\title{Supplementary Materials for AQD: Towards Accurate Quantized Object Detection}

\author{Peng Chen$^2$\thanks{
First two authors contributed equally.} ~~ ~ Jing Liu$^1$\samethanks ~~ ~ Bohan Zhuang$^1$\thanks{
Corresponding author. E-mail: $\tt  bohan.zhuang@monash.edu$}  ~~ ~
Mingkui Tan$^{3}$  ~~ ~ Chunhua Shen$^{1,2}$
\\
	$^1$Monash University ~~ ~
	$^2$University of Adelaide ~~ ~
	$^3$South China University of Technology
}

\maketitle

\renewcommand{\thesection}{S\arabic{section}}
\renewcommand\thefigure{S\arabic{figure}}
\renewcommand{\thetable}{S\arabic{table}}

\section{More Results on ImageNet}
\noindent\textbf{Implementation details.} Following HAQ~\cite{haq2019}, we quantize all the layers, in which the first and the last layers are quantized to 8-bit. Following~\cite{Li2020Additive, Esser2020LEARNED}, we introduce weight normalization during training. We use SGD with nesterov~\cite{nesterov1983method} for optimization, with a momentum of $0.9$. For all models on ImageNet, we first train the full-precision models and then use the pre-trained weights to initialize the quantized models. We then fine-tune for 150 epochs. The learning rate starts at 0.01 and decays with cosine annealing~\cite{loshchilov2016sgdr}.

\noindent\textbf{Main Results.} We apply the proposed method to quantize MobileNetV1~\cite{howard2017mobilenets} and MobileNetV2~\cite{sandler2018inverted} to 4-bit. We compare the performance of different methods in Table~\ref{table:results_on_imagenet}. From the results, our proposed method outperforms other methods by a large margin. For example, compared with HAQ, our proposed method achieve 2.7\% and 3.5\% higher Top-1 accuracy for 4-bit MobileNetV1 and MobileNetV2. 

\begin{table}[!htb]
\caption{Performance comparisons on ImageNet.}
\begin{center}
\scalebox{0.9}
{
\begin{tabular}{cccccccc}
\hline
Network  & Method & \tabincell{c}{Top-1 \\ Acc. (\%)} & \tabincell{c}{Top-5 \\ Acc. (\%)} \\ 
\hline
\multirow{4}{*}{MobileNetV1} & Full-precision & 70.9 & 89.8 \\
& PACT~\cite{choi2018pact} & 62.4 & 84.2 \\
& HAQ~\cite{haq2019} & 67.4 & 87.9 \\
& Ours & 70.1 & 89.3 \\
\hline
\multirow{4}{*}{MobileNetV2} & Full-precision & 71.9 & 90.3 \\
& PACT~\cite{choi2018pact} & 61.4 & 83.7 \\
& HAQ~\cite{haq2019} & 67.0 & 87.3 \\
& Ours & 70.5 & 89.5 \\
\hline
\end{tabular}
}
\end{center}
\label{table:results_on_imagenet}
\end{table}

{\small
\bibliographystyle{ieee_fullname}
\bibliography{reference}
}